\documentclass[letterpaper, 10 pt, conference]{ieeeconf}  

\IEEEoverridecommandlockouts                              

\usepackage{graphics} 
\usepackage{epsfig} 
\usepackage{times} 
\usepackage{amsmath} 
\usepackage{amssymb}  
\usepackage{subfigure}

\title{\LARGE \bf
Unsupervised Learning of 3D Scene Flow from Monocular Camera*
}

\author{Guangming Wang, Xiaoyu Tian, Ruiqi Ding, and Hesheng Wang 
\thanks{*This work was supported in part by the Natural Science Foundation of China under Grant U1613218, U1913204, 62073222, in part by “Shu Guang”project supported by Shanghai Municipal Education Commission and Shanghai Education Development Foundation under Grant 19SG08, in part by grants from NVIDIA Corporation. Corresponding Author: Hesheng Wang. The first two authors contributed equally.
}
\thanks{G. Wang, X. Tian, R. Ding and H. Wang are with Department of Automation, Institute of Medical Robotics, Key Laboratory of System Control and Information Processing of Ministry of Education, Key Laboratory of Marine Intelligent Equipment and System of Ministry of Education, Shanghai Engineering Research Center of Intelligent Control and Management, Shanghai Jiao Tong University, Shanghai 200240, China.}%
}

\begin{document}
\maketitle
\thispagestyle{empty}
\pagestyle{empty}
	
\begin{abstract}

   Scene flow represents the motion of points in the 3D space, which is the counterpart of the optical flow that represents the motion of pixels in the 2D image. However, it is difficult to obtain the ground truth of scene flow in the real scenes, and recent studies are based on synthetic data for training. Therefore, how to train a scene flow network with unsupervised methods based on real-world data shows crucial significance. A novel unsupervised learning method for scene flow is proposed in this paper, which utilizes the images of two consecutive frames taken by monocular camera without the ground truth of scene flow for training. Our method realizes the goal that training scene flow network with real-world data, which bridges the gap between training data and test data and broadens the scope of available data for training. Unsupervised learning of scene flow in this paper mainly consists of two parts: (i) depth estimation and camera pose estimation, and (ii) scene flow estimation based on four different loss functions. Depth estimation and camera pose estimation obtain the depth maps and camera pose between two consecutive frames, which provide further information for the next scene flow estimation. After that, we used depth consistency loss, dynamic-static consistency loss, Chamfer loss, and Laplacian regularization loss to carry out unsupervised training of the scene flow network. To our knowledge, this is the first paper that realizes the unsupervised learning of 3D scene flow from monocular camera. The experiment results on KITTI show that our method for unsupervised learning of scene flow meets great performance compared to traditional methods Iterative Closest Point (ICP) and Fast Global Registration (FGR). The source code is available at: https://github.com/IRMVLab/3DUnMonoFlow.

\end{abstract}

\section{Introduction}

The 3D perception of the scene is crucial to the upper-level control and planning of robots \cite{liu2016formation}. Scene flow represents the vector that describes the motion of a point in 3D space from the position in the first frame to the position in the second frame. Optical flow is its counterpart if scene flow gets projected to 2D image plane \cite{vedula1999three}. Motion estimation of each point in 3D space can provide rich information for motion segmentation, object localization \cite{zhao2019self}, pose estimation \cite{liu2019local}, target tracking \cite{shi2016visual}, etc. For non-deep learning methods, variational methods are widely used to estimate scene flow \cite{vogel20153d,menze2015object,menze2018object,lv2016continuous,huguet2007variational}. With the development of deep learning \cite{hussain2019depth}, the scene flow is estimated by combining disparity map \cite{wang2019depth} and optical flow \cite{mayer2016large,lv2018learning,jiang2019sense}. 

\begin{figure}[t]
   \centering
   \includegraphics[width=1.00\linewidth]{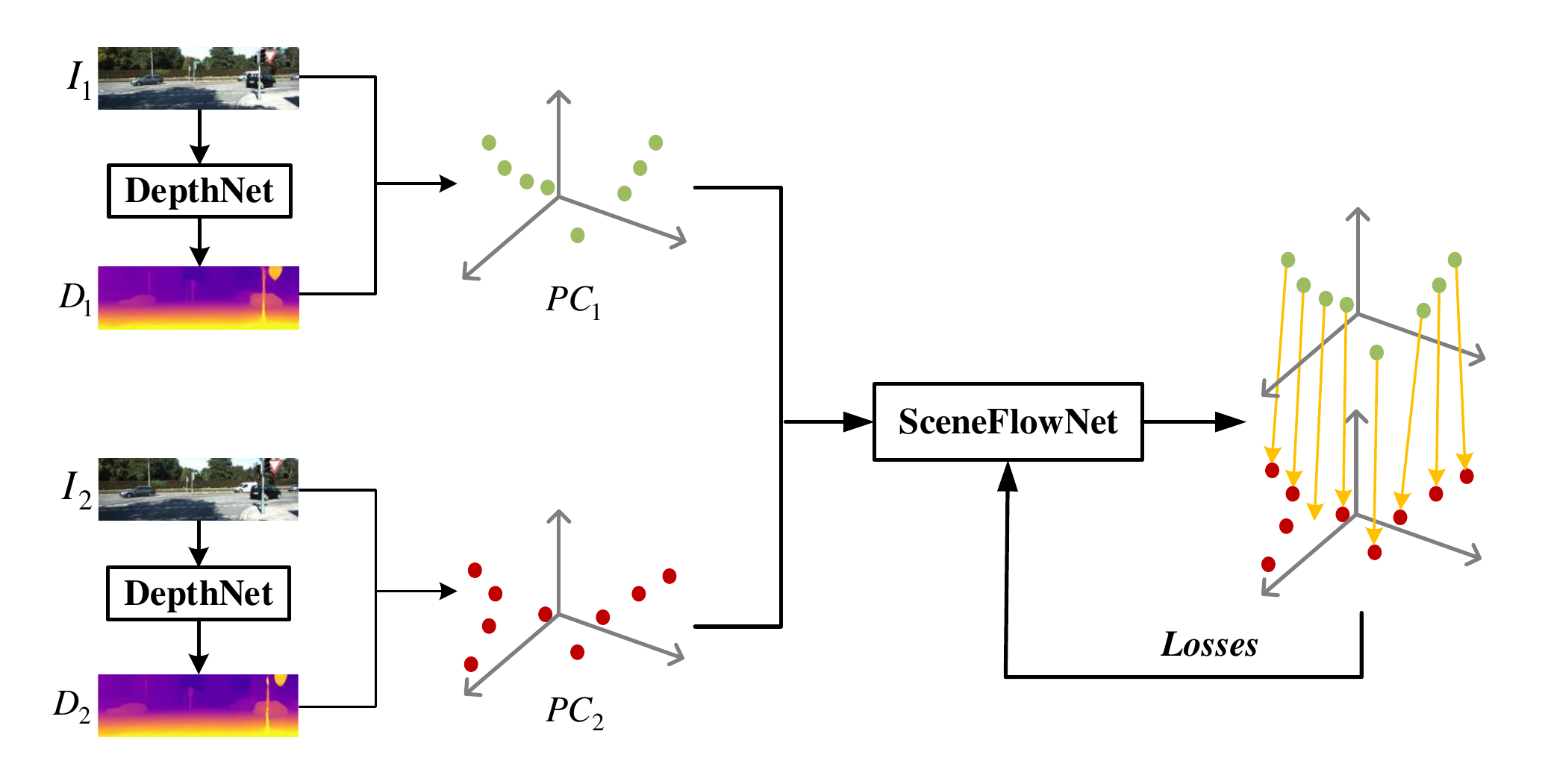}
   \caption{Unsupervised learning of 3D scene flow from monocular camera. Our model do not use the ground truth flow during training. The "SceneFlowNet" can be any network for end-to-end scene flow estimation.}
   \label{figure:pointcloud}
\end{figure}

With the development of deep learning methods for point clouds, estimation methods for scene flow based on point clouds are also developed recently \cite{gu2019hplflownet,liu2019flownet3d}. Scene flow network can provide the scene flow between the two given point clouds which are corresponding to two consecutive frames. The network can implement end-to-end estimation of scene flow and has been realized by \cite{gu2019hplflownet,liu2019flownet3d} recently. The problem solved by scene flow network is described as: with the inputs of the point cloud in the first frame $PC_{1}$ and the point cloud in the second frame $PC_{2}$, the scene flow network can estimate the vector $sf_i=(dx_i,dy_i,dz_i)$ for each point in the first frame that describes how the point moves to the location in the second frame. Due to the inhomogeneity during the sampling of point clouds, it is possible that for a point in the first frame there is no corresponding point in the second frame. In other words, a point in the first frame might not be able to match any point in the second frame after the movement via scene flow $sf_i$. The character is very similar to optical flow, which enhances the difficulty of the acquisition of ground truth for optical flow and scene flow. Most supervised research on optical flow has been done on synthetic datasets \cite{dosovitskiy2015flownet}, as well as scene flow  \cite{gu2019hplflownet,liu2019flownet3d}. That is why unsupervised learning is as important to scene flow as optical flow \cite{ren2017unsupervised,meister2018unflow,wang2018occlusion,janai2018unsupervised,liu2019ddflow,liu2019selflow}. The real-world image data can be used in the unsupervised training of scene flow with suitable losses as shown in Fig.~\ref{figure:pointcloud}, which bridges the gap between training data and testing data \cite{menze2018object}. This paper is devoted to this.

Our contributions are as follows:
\begin{itemize}

\item The point clouds of two consecutive frames are obtained from 2D images and depth maps,  while depth maps can be acquired through inputting the images into the depth estimation network. Depth consistency  of scene flow is proposed as penalty loss for the unsupervised learning of scene flow inspired by unsupervised learning of depth and pose estimation. 
\item The concepts of dynamic scene flow and static scene flow are presented in this paper. Static scene flow can be obtained from camera pose transformation. 
\item The loss for dynamic-static consistency  is proposed. Overall scene flow, dynamic scene flow, and static scene flow which are generated from point clouds should satisfy a consistency relationship in geometry. The violation of consistency is punished for unsupervised learning of scene flow.
\item  The experiment was conducted on KITTI Odometry dataset \cite{geiger2013vision} and KITTI scene flow dataset \cite{menze2018object}, in which the odometry dataset was used for unsupervised training and the scene flow dataset was used for testing. The results showed that our unsupervised method could achieve super performance than some traditional methods.

\end{itemize}

\section{Related Work}
\subsection{Scene Flow Estimation}

Scene flow can be obtained from a variety of sources.   Menze et al. \cite{menze2015object} divided the objects of the scene into independent moving rigid objects, and two consecutive frames of binocular images were used to estimate scene flow with discrete-continuous Conditional Random Field (CRF). Bounding boxes, segmentation, and object recognition  are introduced in \cite{behl2017bounding}. Then in \cite{menze2018object}, the robust discrete optical flow was combined to track large displacement.   Lv et al. \cite{lv2018learning}  proposed the joint supervised learning of camera motion and mask for rigid parts with the synthesized RGB-D dataset, and the scene flow is estimated by optical flow in non-rigid regions and static flow. Battrawy et al. \cite{battrawy2019lidar} fuse sparse LiDAR and stereo images to estimate the scene flow. With the seeds built from LiDAR, the matches of stereo images become more robust and the mismatches can be filtered after the consistency check.  Later, optical flow, segmentation, and parallax estimation based on deep learning are combined as an energy minimization problem for scene flow estimation  \cite{ma2019deep}. Jiang et al. \cite{jiang2019sense} proposed a novel network to estimate scene flow from stereo which is more compact by sharing features with a shared encoder.

Scene flow can be applied to point clouds directly and the estimation of scene flow with point clouds is underexplored. Dewan et al. \cite{dewan2016rigid} reduced the scene flow with point cloud to an equivalent energy minimization problem and the analysis of dynamics is based on point level. Ushani et al. \cite{ushani2017learning} obtained scene flow from a filtered occupancy grid which is constructed by raw LiDAR data and then the foreground is extracted. With the development of deep learning for 3D point clouds \cite{qi2017pointnet,qi2017pointnet++,liu2019lpd,wang2020spherical,wang2020anchor}, FlowNet3D \cite{liu2019flownet3d} was proposed and could directly process point clouds and generate 3D scene flow from point clouds in the end-to-end style. Behl et al. \cite{behl2019pointflownet} proposed a trainable model for joint scene flow estimation and rigid motion prediction from unstructured LiDAR data. A learnable operator named parametric continuous convolution which can operate over non-rigid structured data was proposed by Wang et al. \cite{wang2018deep}.  HPLFlowNet \cite{gu2019hplflownet} proposed DownBCL, UpBCL, and CorrBCL operations to restore the rich information in point clouds. Wu et al. \cite{wu2020pointpwc} introduced a learnable cost volume layer and proposed PointPWC-Net to directly process point clouds in a coarse-to-fine fashion. A hierarchical neural network with a double attentive embedding layer was proposed in \cite{wang2020hierarchical} for the scene flow estimation of two consecutive point clouds.

\subsection{Unsupervised Estimation of Optical Flow}
Scene flow based on deep learning with the input of point clouds directly is just developing \cite{liu2019flownet3d,gu2019hplflownet,wu2020pointpwc,wang2020hierarchical}. To our knowledge, there is no unsupervised learning for 3D scene flow from only monocular videos by the time the paper was finished. Since scene flow and optical flow are very similar, unsupervised learning for optical flow is reviewed here. The movement of pixels in the image is defined as optical flow which is a 2D vector. Dosovitskiy et al. \cite{dosovitskiy2015flownet} firstly proposed a convolutional network to estimate the optical flow with a supervised method. The structure of the network gets improved later \cite{ilg2017flownet}. Early networks are trained on the synthetic dataset \cite{mayer2016large}, that is because the ground truth of optical flow in the real world is difficult to obtain, which increases the difficulty for supervision and estimation. As a result, the method of unsupervised learning on real-world data has attracted much attention \cite{meister2018unflow,besl1992method,wang2018occlusion,janai2018unsupervised,liu2019ddflow,liu2019selflow}.

Ren et al. \cite{ren2017unsupervised} used image warping to minimize photometric consistency in unsupervised training and estimating for optical flow. Meister et al. \cite{meister2018unflow} introduced consistency loss and a method with empirical parameters for occlusion in the process of computing consistency and warping images. Occlusion is considered as an important issue in the estimation of disparity, optical flow, and scene flow in recent years \cite{ilg2018occlusions,wang2018occlusion,janai2018unsupervised,liu2019ddflow,liu2019selflow}. More recently, works have continued to address this problem by distilling the knowledge of existing networks to improve the accuracy of unsupervised estimation  for optical flow \cite{liu2019ddflow,liu2019selflow}. Recently, some works focus on unsupervised learning of optical flow combined with depth and pose \cite{yin2018geonet,Ranjan2019CCNet,wang2020unsupervised}. GeoNet \cite{yin2018geonet} is proposed to jointly estimate monocular depth, optical flow, and ego-motion from videos and the optical flow learns the residual motion of scenes. Ranjan et al. \cite{Ranjan2019CCNet} introduced a dynamic and static segmentation based framework to coordinate the training of various specialized networks. Similarly, pixels are divided into rigid regions, non-rigid regions, and occluded regions in \cite{wang2020unsupervised, wang2019unsupervised} to train several networks.  Hur et al. \cite{hur2020self} proposed to estimate scene flow based on images, but stereo videos are used for training, while our method only needs monocular videos to learn scene flow estimation. This paper is inspired by the above.

\section{Unsupervised Learning of Scene Flow}
\label{section:unsupervised learning}
\subsection{Problem Description}
\label{section:Problem description}

\begin{figure*}[t]
   \centering
   \includegraphics[width=1.00\linewidth]{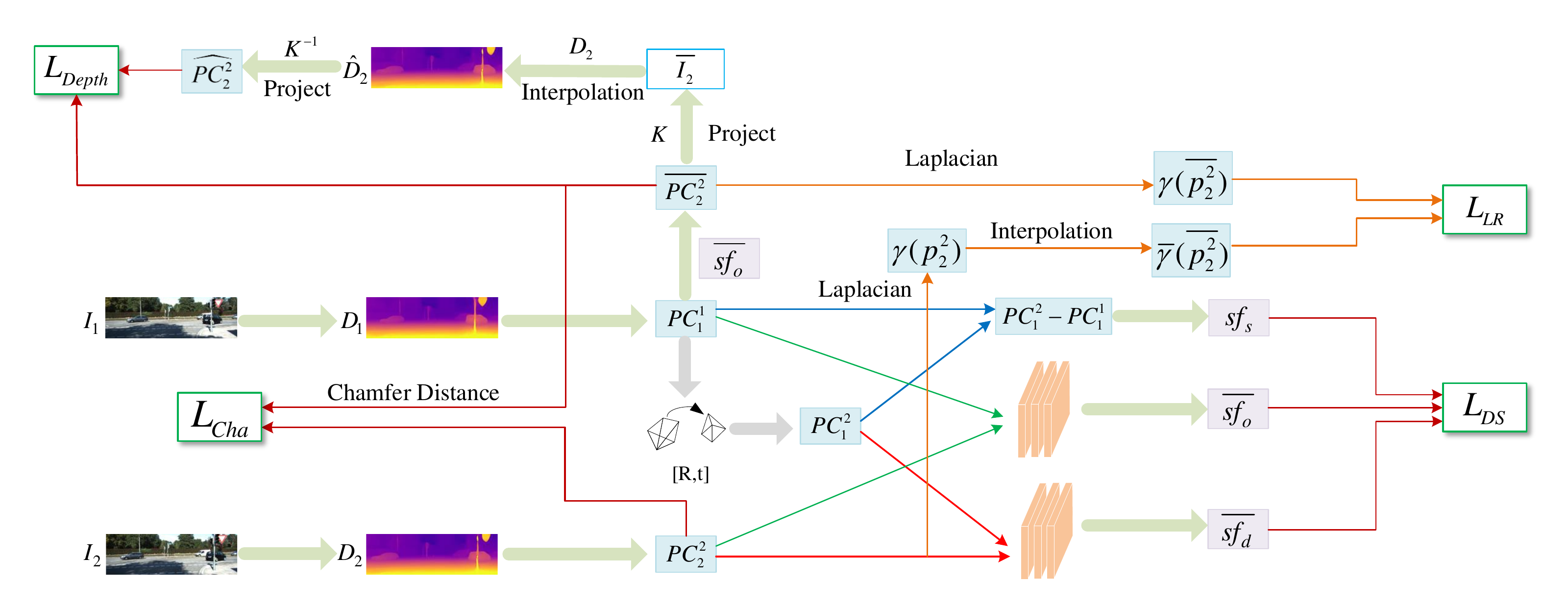}
   \vspace{-0.6cm}
   \caption{Four different loss functions are used in this paper. $L_{Depth}$ represents the distance between the predicted second point cloud $\overline{PC_2^2}$ and its interpolated counterpart. $L_{DS}$ evaluates the consistency between overall flow and the sum of static flow and dynamic flow. $L_{Cha}$ represents Chamfer distance between $PC^2_2$ and $\overline{PC_2^2}$. $L_{LR}$ represents the difference between the Laplacian of $\overline{PC^2_2}$ and its interpolated counterpart. }   
   \label{figure:losses}
\end{figure*}

Given two consecutive frames of images $I_1$ and $I_2$ for the same scene, we can get the depth estimation network and pose network by the unsupervised learning \cite{bian2019unsupervised}. Then, the depth maps of two images are obtained through the depth estimation network. From the images and depth maps, two consecutive frames of point clouds can be obtained. Further, what we need to get is the 3D scene flow transformation $sf$  without the constraint between each point to the other in a frame. With the ideal scene flow $sf$, the physical position of the spatial point corresponding to the first frame can be obtained in the second frame.

By inputting two consecutive images into the depth network respectively, the corresponding depth maps $D_1$ and $D_2$ can be obtained. The point clouds $X = (PC_1^1,F_1)$ and $Y=(PC_2^2,F_2)$ corresponding to the pixels of the above two frames can be easily obtained from the depth maps of the two frames with the camera intrinsic matrix $K$, where $PC_i^i$ is the points in Cartesian coordinates obtained from: 
\begin{equation}
    \label{eq:depth-pc}
     PC_i^i = D_i \cdot K^{-1} \cdot I_i, i\in\{1,2\},
\end{equation}
in which each pixel $p_{u,v}=(u,v,1)$ in image $I_i$ represents the pixel coordinate on the image plane. $D_i$ represents the depth map. $F_i=(R_i,G_i,B_i)$   $(i\in\{1,2\})$ represents the color features of the point clouds. With the point clouds of the two images, the scene flow $sf$ corresponding to each point in the first frame is desired. However, this is not an easy task. On the one hand, similar to optical flow, not all points in the first frame can be found in the second frame due to occlusion. On the other hand, when scenes in the real world are sampled, the points in $PC_1^1$ not always coincide with the points in $PC_2^2$. All these problems bring difficulties to our unsupervised scene flow learning.

In order to perform training on the unlabeled image datasets, four loss functions are adopted: (1) depth consistency loss, (2) dynamic-static consistency loss, (3) Chamfer loss, and (4) Laplacian regularization loss. In the meanwhile, we apply the loss functions to each level of the predicted scene flow so as to make full use of the hierarchical structure of the scene flow network \cite{wang2020hierarchical}. The specific loss functions are described separately in the following sections.

\subsection{Depth Consistency Loss}
\label{section:Depth Consistency}

For the unlabeled datasets, the scene flow network cannot be trained directly by minimizing the difference between ground truth flow and the estimated flow due to the lack of annotated ground truth flow. To solve the problem, local consistency is used to design depth consistency loss. First, point clouds $PC_1^1$ and $PC_2^2$ of the first and the second frames are input into the scene flow network, and the estimated scene flow $\overline{sf_o}$ can be obtained. Let $\overline {PC_2^2}  = PC_1^1 + \overline {sf_o}$ be the position where point cloud $PC_1^1$ of the first frame flow to the second frame and $\overline{D_2}$ be the depth values of $\overline {PC_2^2}$. With the projection equation:
\begin{equation}
    \label{projection-depth}
     \bar I_2 = \frac{1}{\overline{D_2}} \cdot K \cdot \overline{PC_2^2},
\end{equation}
 $\overline{PC_2^2}$ can be projected on the image plane of the second frame. $(u,v)$ represents the projected pixel coordinate in $\bar I_2$ for each point of $\overline{PC_2^2}$. And then the interpolated depth map ${{\hat D}_2}$ is acquired through the following bilinear interpolation:
\begin{equation}
\begin{aligned}
\label{interpolation-depth}
    \hat D_2(u,v) & = \frac{{D_2({i_1},{j_1})}}{{({i_2} - {i_1})({j_2} - {j_1})}}({i_2} - u)({j_2} - v) \\
    &+ \frac{{D_2({i_2},{j_1})}}{{({i_2} - {i_1})({j_2} - {j_1})}}(u - {i_1})({j_2} - v) + \\
    &\frac{{D_2({i_1},{j_2})}}{{({i_2} - {i_1})({j_2} - {j_1})}}({i_2} - u)(v - {j_1}) + \\
    &\frac{{D_2({i_2},{j_2})}}{{({i_2} - {i_1})({j_2} - {j_1})}}(u - {i_1})(v - {j_1}).
\end{aligned}
\end{equation}
The four points $({i_m},{j_n}),(m,n\in\{1,2\})$ represent the four nearest points of $(u, v)$ on the depth map. $D_2({i_m},{j_n}),(m,n\in\{1,2\})$ represents the depth values of the four points. $\hat D_2(u,v)$ represents the depth value of point $(u, v)$.
 Then, through the projection of $\bar I_2$ with the new depth map ${{\hat D}_2}$ into 3D space,
\begin{equation}
    \label{eq:depth-pc2}
     \widehat {PC_2^2} = {\hat D}_2 \cdot K^{-1} \cdot \bar I_2,
\end{equation}
the point cloud $\widehat {PC_2^2}$ of the second frame after interpolation can be obtained. If the scene flow estimation is accurate enough, then point clouds $\overline {PC_2^2}$ and $\widehat {PC_2^2}$ should be in the same physical position. Therefore, depth consistency loss is defined as the Euclidean distance between these points:
\begin{equation}
    \label{eq:depth-consist}
    L_{Depth} = \sum ||\overline {PC_2^2}  - \widehat {PC_2^2}|{|^2}.
\end{equation}

\subsection{Dynamic-static Consistency Loss}
\label{section:Dynamic-static Consistency}
The scene flow of the point cloud from the first frame to the second frame is generated by the transformation of the camera ego-motion and the movement of objects in the scene. Therefore, the sum of the scene flow generated by the transformation of camera pose and the scene flow generated by the movement of objects in the scene is consistent with the overall scene flow, which is called dynamic-static consistency in this paper. Images $I_1$ and $I_2$ of the first and the second frames are input into the pose network obtained by the joint unsupervised learning of depth and pose, and the transformation of camera pose $(R,t)$ from the first frame to the second frame can be obtained. $R$ represents rotation matrix and $t$ represents translation vector. Based on this, the position $PC_1^2$ of  the point cloud $\bar X = (PC_1^2,{F_1})$, which represents the coordinates of the point cloud $X$ observed in the second frame, can be obtained from the transformation:
\begin{equation}
    \label{eq:trans-pose}
    PC_1^2 = PC_1^1 \cdot R + t.
\end{equation}
 As shown in Fig.~\ref{figure:losses}, $sf_s = PC_1^2 - PC_1^1$ is defined as static flow from the first frame to the second frame. Then, $\bar X$ and $Y$ are input into the scene flow network to obtain the estimated dynamic flow $\overline{sf_d}$, i.e. the scene flow generated by moving objects. According to dynamic-static consistency, if the scene flow estimation is accurate, the overall flow from the point cloud of the first frame to the second frame should satisfy $\overline{sf_o}= sf_s + \overline{sf_d}$. Hence, the difference between the overall flow and the sum of static flow and dynamic flow forms dynamic-static consistency loss:
\begin{equation}
    \label{eq:ds-consist}
    L_{DS} = \sum ||\overline {sf_o} - (sf_s + \overline {sf_d} )|{|^2}.
\end{equation}

\begin{figure*}[t]
	\begin{center}
		\resizebox{1.0\textwidth}{!}
		{
			\setlength{\tabcolsep}{0.9mm}
		\begin{tabular}{cccc}
			\subfigure{\includegraphics[width=0.25\linewidth]{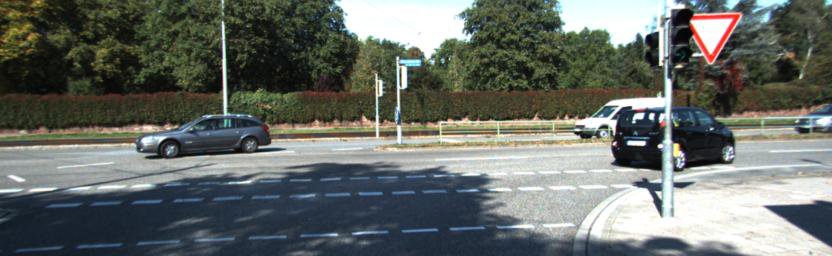}}
			&\subfigure{\includegraphics[width=0.25\linewidth]{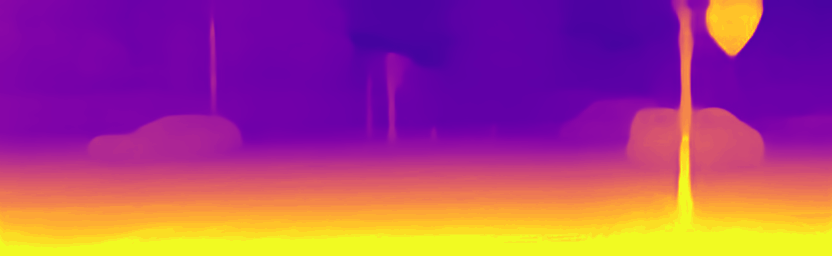}}&\includegraphics[width=0.25\linewidth]{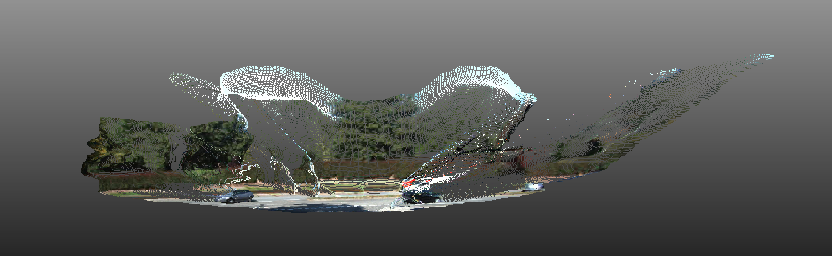}&\includegraphics[width=0.25\linewidth]{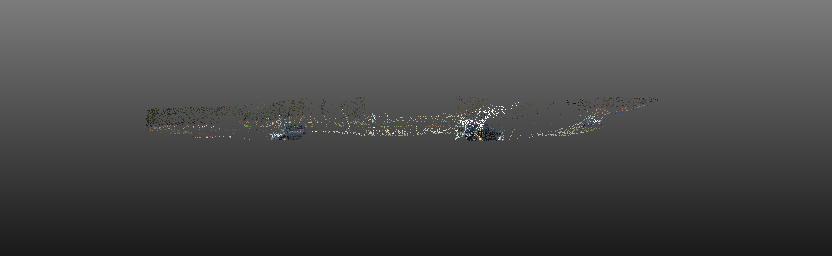}
		\\	\subfigure{\includegraphics[width=0.25\linewidth]{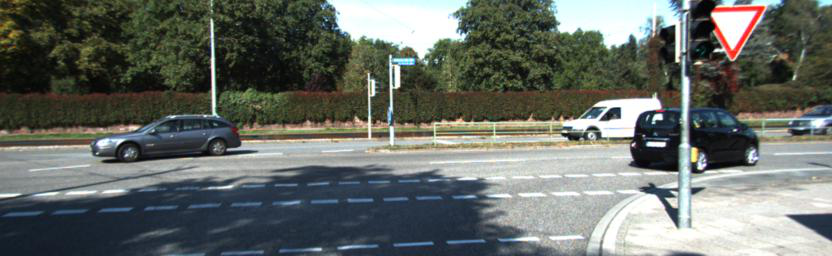}}
		&\subfigure{\includegraphics[width=0.25\linewidth]{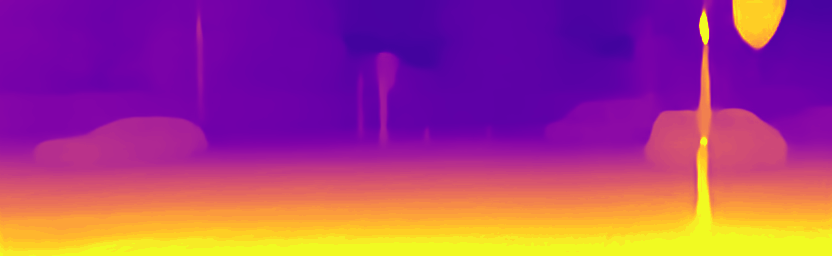}}&\includegraphics[width=0.25\linewidth]{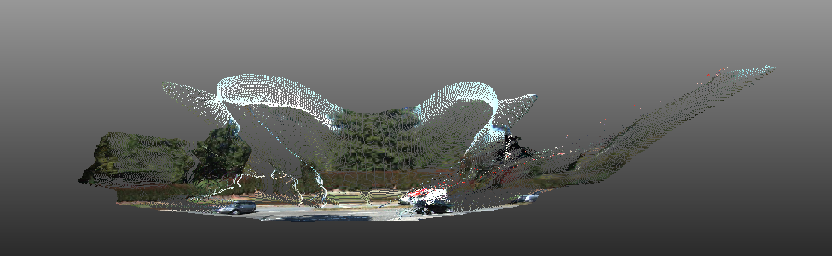}&\includegraphics[width=0.25\linewidth]{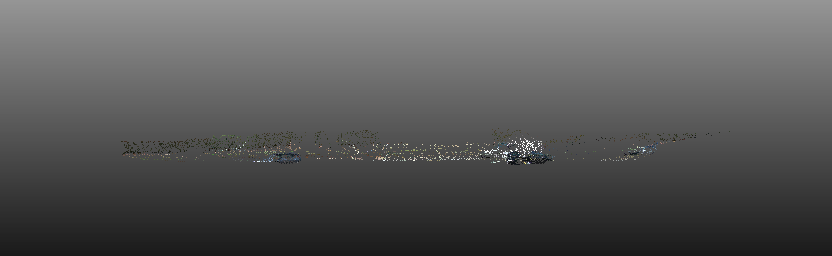}
		\vspace{0.3cm}\\		
		
		\subfigure{\includegraphics[width=0.25\linewidth]{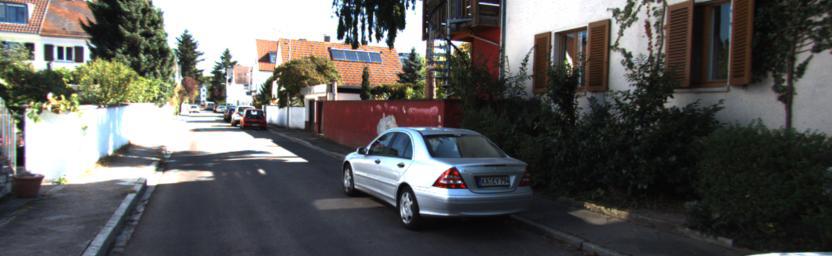}}
		&\subfigure{\includegraphics[width=0.25\linewidth]{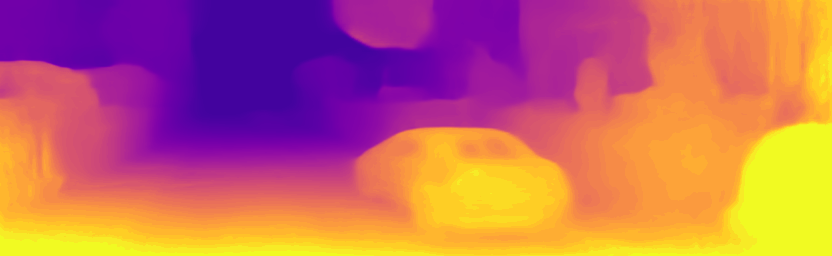}}&\includegraphics[width=0.25\linewidth]{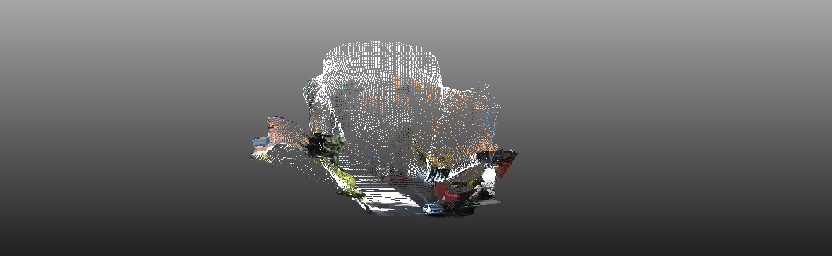}&\includegraphics[width=0.25\linewidth]{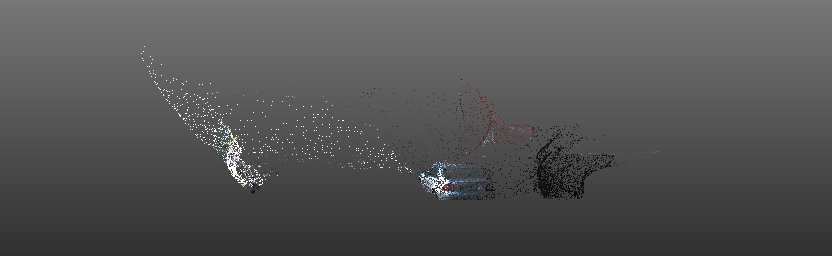}
		\\	\subfigure{\includegraphics[width=0.25\linewidth]{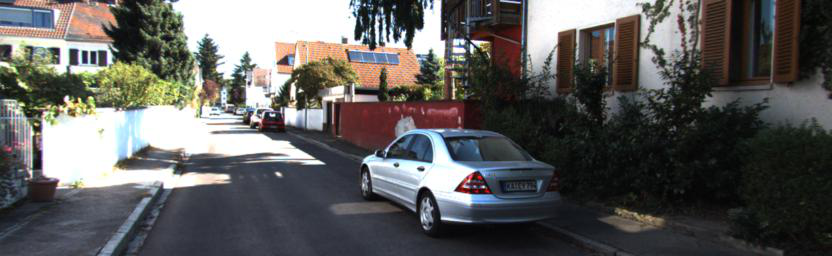}}
		&\subfigure{\includegraphics[width=0.25\linewidth]{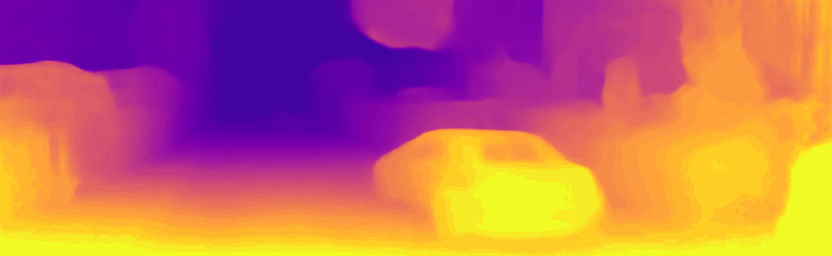}}&\includegraphics[width=0.25\linewidth]{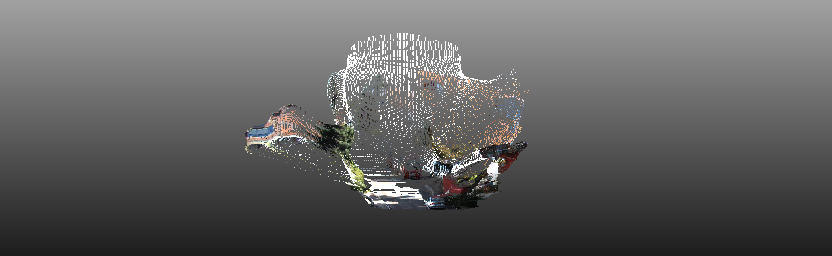}&\includegraphics[width=0.25\linewidth]{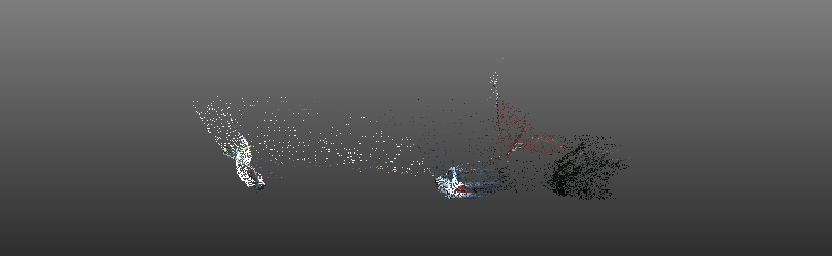}
		\vspace{0.3cm}\\		
		
		\subfigure{\includegraphics[width=0.25\linewidth]{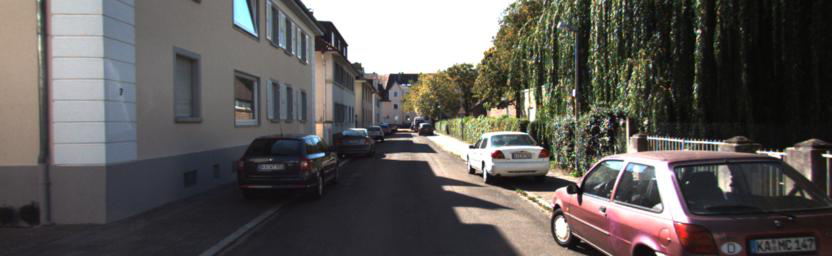}}
		&\subfigure{\includegraphics[width=0.25\linewidth]{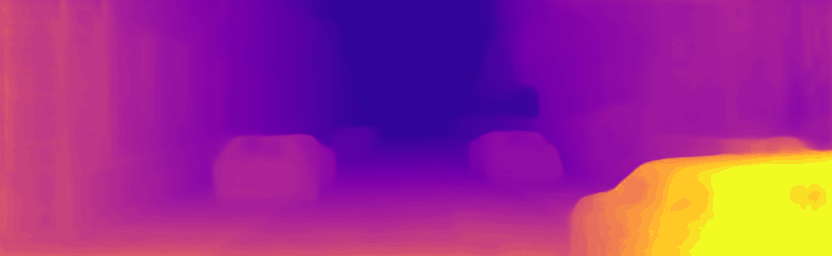}}&\includegraphics[width=0.25\linewidth]{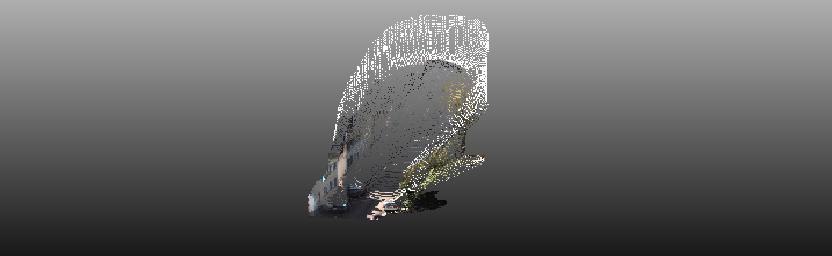}&\includegraphics[width=0.25\linewidth]{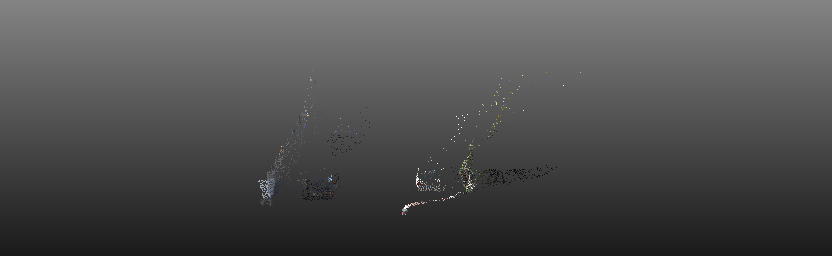}
		\\	\subfigure{\includegraphics[width=0.25\linewidth]{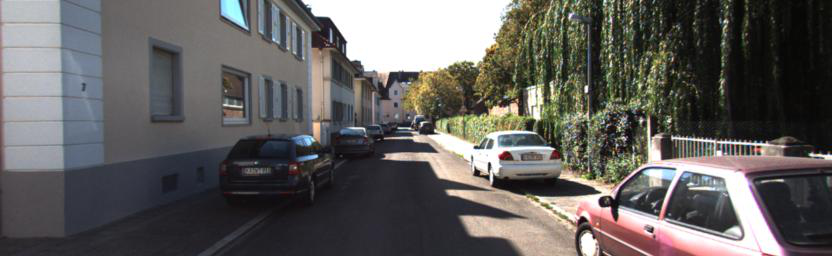}}
		&\subfigure{\includegraphics[width=0.25\linewidth]{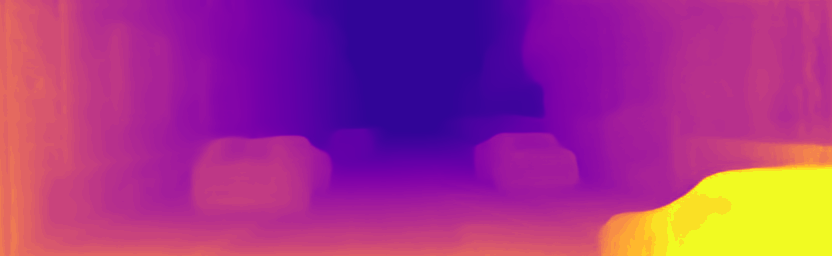}}&\includegraphics[width=0.25\linewidth]{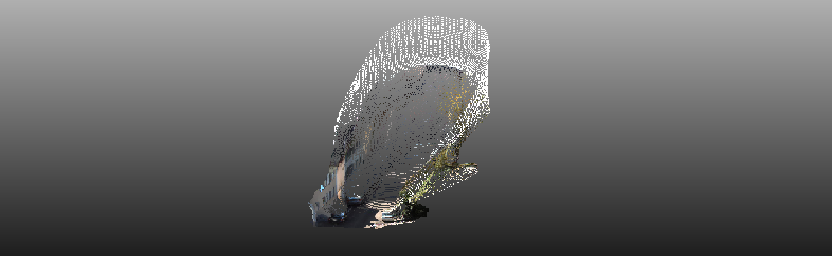}&\includegraphics[width=0.25\linewidth]{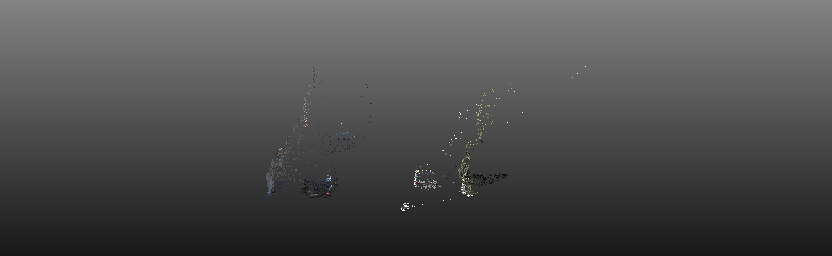}

		\\{\small (a) RGB image} &  {\small (b) Depth map}&{\small (c) Original point cloud}&{\small (d) Point cloud after removal}

		\end{tabular}
	}
	\end{center}
	\caption{\label{fig:visual} Several sets of visualization results for two consecutive frames. The two frames next to each other are consecutive frames.}
\end{figure*}

\subsection{Chamfer Loss}
\label{section:Chamfer}
In order to make the point cloud of the first frame after scene flow warping close to the point cloud of the second frame as much as possible, we choose Chamfer distance as one of our loss functions as in \cite{wu2020pointpwc}.
As defined in Section~\ref{section:Depth Consistency}, $\overline{PC^2_2}$ represents the point cloud warped from the  point cloud of the first frame according to the predicted scene flow. Therefore, the Chamfer loss can be described as:
\begin{small}
\begin{equation}
    \label{eq:Chamfer}
    L_{Cha}=\sum_{\overline{p^2_2} \in \overline{PC^2_2}} \min _{p^2_2 \in PC^2_2}\left\|\overline{p^2_2}-p^2_2\right\|^{2}+\sum_{p^2_2 \in PC^2_2} \min _{\overline{p^2_2} \in \overline{PC^2_2}}\left\|\overline{p^2_2}-p^2_2\right\|^{2},
\end{equation}
\end{small}
where $\overline{p^2_2}$ is the point of $\overline{PC^2_2}$ and $p^2_2$ is the point of $PC^2_2$.

\subsection{Laplacian Regularization Loss}
\label{section:Laplacian}
We refer to previous works \cite{wu2020pointpwc,sorkine2005laplacian} and use Laplacian regularization as our loss function. Assuming that the predicted scene flow is accurate, the Laplacian of the warped point cloud  $\overline{PC^2_2}$ should be the same as the Laplacian of point cloud $PC^2_2$. Let $N(p_i)$ be a local area around point $p_i$ and $|N(p_i)|$ be the number of points in $N(p_i)$. The Laplacian coordinate vector $\gamma\left(p_i\right)$ of $p_i$ can be represented as:
\begin{equation}
    \label{eq:LapCoordVec}
    \gamma\left(p_i\right)=\frac{1}{\left|N\left(p_i\right)\right|} \sum_{p_j \in N\left(p_i\right)}\left(p_j-p_i\right),
\end{equation}
and the Laplacian of $\overline{PC^2_2}$ and $PC^2_2$ can be computed. Then we can obtain the interpolated Laplacian coordinate vector $\overline\gamma(\overline{p^2_{2}})$ of $\overline{PC^2_2}$  through inverse distance weighted interpolation:
\begin{equation}
    \label{eq:invinter}
    \overline\gamma\left(\overline{p^2_{2}}\right)=\frac{\sum_{j=1}^{k} w\left(\overline{p^2_2}, p_j\right) \gamma\left(p_j\right)}{\sum_{j=1}^{k} w\left(\overline{p^2_2}, p_j\right)},
\end{equation}
in which $w\left(\overline{p^2_2}, p_j\right)=1 / d\left(\overline{p^2_2}, p_j\right)$, $p_j\in N(\overline{p^2_2})\cup {{PC^2_2}}$,  $d\left(\overline{p^2_2}, p_j\right)$ is the Euclidean distance between two points. Hence, the Laplacian regularization loss is described as:
\begin{equation}
    \label{eq:Laplacian}
    L_{LR}=\sum_{\overline{p^2_2}\in \overline{PC^2_2}}\left\|\gamma\left(\overline{p^2_2}\right)-\overline\gamma\left(\overline{p^2_{2}}\right)\right\|^{2}.
\end{equation}

\subsection{Unsupervised Loss}
As described in section~\ref{section:Dynamic-static Consistency}, the overall scene flow can be replaced by the sum of dynamic scene flow and static scene flow. Therefore,  $\overline{PC^2_2}$ in section~\ref{section:Depth Consistency} can be presented in two ways, and depth consistency loss has two different ways to compute. Let $L_{Depth1}$ be the loss computed from the overall scene flow, $L_{Depth2}$ be the loss computed from the sum of dynamic and static scene flow, and $L_{Depth}=L_{Depth1}+L_{Depth2}$.
The overall unsupervised loss is a weighted sum of all loss functions  over all levels of scene flow estimation:
\begin{small}
\begin{equation}
    \label{eq:unsupervised}
    L_{Un}=\sum_{k=1}^{K} \alpha_{k}\left(\beta_{1} L^k_{Depth}+\beta_{2} L^k_{DS}+\beta_{3} L^k_{Cha}+\beta_{4}L^k_{LR}\right),
\end{equation}
\end{small}
where $\alpha_k$ are factors for different levels of scene flow outputs and $\beta_i$ are weights for different loss functions.

\section{Experiments}

\subsection{Datasets and Data Preprocessing}

As described in section ~\ref{section:unsupervised learning}, our data contains colorized point clouds, therefore we can input colors as features into the scene flow network. The models are trained with our approach on KITTI Odometry dataset \cite{geiger2013vision} and tested on KITTI scene flow dataset \cite{menze2018object}. KITTI Odometry dataset is a dataset of image sequences of consecutive frames, which is stored in PNG format and contains 22 sequences. The dataset is widely used in odometry estimation and includes vehicles, pedestrians, and other complex objects in various situations like city, road, campus, etc.

With the depth values of the images which are obtained from the depth network, the pixels in the first frame can be projected into the camera coordinate system of the first frame and represented as $PC_1^1$. In a similar way, the pixels in the second frame can be projected into the camera coordinate system of the second frame and represented as  $PC_2^2$. For each frame of images in the dataset, we can obtain the point cloud corresponding to the pixels in the 3D space.


\begin{table*}[t]
\centering
\caption{Evaluation results on the KITTI scene flow dataset \cite{menze2015object}.}
\begin{tabular}{lcccccc}
\hline
\multicolumn{1}{c}{Method}       & EPE3D & Acc3D Strict & Acc3D Relax & Outliers3D \\  \hline
ICP \cite{besl1992method}          & 0.5181     & 0.0669 & 0.1667  & 0.8712 \\
FGR \cite{zhou2016fast}            & 0.4835      & 0.1331  & 0.2851 & 0.7761 \\
Ours                               &  0.4470    &  0.1470 &  0.3521 & 0.7739 \\ \hline
\end{tabular}

\label{table:Results}
   
\end{table*}


\begin{table*}[ht]
   \centering
   \caption{Ablation study on different loss functions  evaluated on the KITTI scene flow dataset \cite{menze2015object}. "Depth" means depth consistency loss.  "DSC" means dynamic-static consistency loss. "Cha" means Chamfer loss. "LR" means Laplacian regularization loss.}
   \begin{tabular}{lcccccc}
   \hline
   \multicolumn{1}{c}{Method}      & EPE3D   &  Acc3D Strict   &   Acc3D Relax   & Outliers3D   \\ \hline
   DSC + Cha + LR      & 0.4834  &  0.1336  &  0.3168 & 0.7904  \\
   Depth + Cha + LR    & 0.5644  &  0.0749  &  0.2187 & 0.8742 \\
   Depth + DSC + LR        & 5.7601  &  0.0184  &  0.0851 & 0.9433 \\
   Depth + DSC + Cha          & 0.8587  &  0.1059  &  0.2628 &  0.8345 \\
   Depth + DSC + Cha + LR & 0.4470 & 0.1470 & 0.3521 & 0.7739 \\  \hline
   \end{tabular}

\label{table:ablation1}
\end{table*}


\begin{table*}[ht]
   \centering
   \caption{Ablation study about different point cloud processing methods during training evaluated on the KITTI scene flow dataset \cite{menze2015object}. "With ground" means that only sky points are removed. In the same way, "With sky" means that only ground points are removed.}
   \begin{tabular}{lccccc}
   \hline
   \multicolumn{1}{c}{Method}      & EPE3D   & Acc3D Strict & Acc3D Relax  & Outliers3D  \\ \hline
   With ground                     &  0.4927  &  0.0928     &  0.2640      & 0.8294 \\
   With sky                        &  0.4477  &  0.0840     &  0.2693       & 0.8322 \\
   Without ground or sky           & 0.4470   & 0.1470      & 0.3521        & 0.7739 \\  \hline
   \end{tabular}

\label{table:ablation2}
\end{table*}

Our experiment is based on the point clouds of two consecutive frames without the need for the ground truth of scene flow. Since the movement of the points from the ground and sky has no practical significance for the scene flow estimation, and too many sky and ground points may even have a negative impact on the subsequent sampling, we removed the points of the ground and sky from our training data. To be specific, we define points beyond a certain range as either ground or sky, and we removed these points from the point cloud of a frame. In this paper, we choose $1.15m$ as the height range and $35m$ as the depth range. Points whose height is below $1.15m$ and whose depth is beyond $35m$ are regarded as the ground and the sky respectively. In the meanwhile, because 3D LiDAR has a certain scanning range, we removed the points out of the range of LiDAR. Also, we compared the depth maps of the two frames and removed occluded regions. The point cloud visualization before and after removal is shown in Fig.~\ref{fig:visual}.

KITTI scene flow dataset contains 200 scenes for training and 200 scenes for test which is generated by establishing a 3D CAD model for the dynamic objects. Liu et al. \cite{liu2019flownet3d} processed the first 150 data points from KITTI scene flow dataset and removed the ground points. Each of the dataset contain two frames of point clouds and the ground truth scene flow of the first frame derived from the optical flow.

\subsection{Network Settings}

The scene flow network in \cite{wang2020hierarchical} is adopted in this paper, which can estimate the scene flow between two point clouds as an end-to-end process. This network has a more-for-less hierarchical architecture, which means that there are more input points than output points. Since the joint learning of depth and pose from monocular videos have achieved good results, we chose the depth network and camera motion network  proposed in \cite{bian2019unsupervised} as our networks.

The network is trained under the guidance of minimizing loss $L_{Un}$. The factors are set as $\alpha_1=0.02, \alpha_2=0.04, \alpha_3=0.08$, and $\alpha_4=0.16$. The weight parameters are set as $\beta_1=0.1, \beta_2=0.1, \beta_3=1.0,$ and $\beta_4=0.3$.
Adam optimization is adopted in our training and the initial learning rate is set as $10^{-3}$. Our experiments are implemented on a Titan RTX GPU with PyTorch 1.7.1, and the training batch size is 5.

\subsection{Evaluation Criteria}

The evaluation method on 3D scene flow in \cite{wang2020hierarchical} is followed. The evaluation criteria included: EPE3D: end point  error in 3D, whose value is the L-2 norm of the difference between the predicted flow and the ground truth flow for each point after averaging. Acc3D Strict: the accuracy for scene flow which represents the percentage of points with EPE3D less than $0.05m$ or relative error less than 5$\%$. Acc3D Relax: the accuracy for scene flow which represents the percentage of points with EPE3D less than $0.1m$ or relative error less than 10$\%$. Outliers3D: the percentage of points with EPE3D more than $0.3m$ or relative error more than 10$\%$.

\subsection{Results}
Quantitative results are shown in Table~\ref{table:Results}. We implemented the unsupervised scene flow training based only on monocular images and the scene flow network trained with our unsupervised methods performs well on the evaluation criteria compared to traditional methods Iterative Closest Point (ICP) \cite{besl1992method} and Fast Global Registration (FGR) \cite{zhou2016fast}.

\subsection{Ablation Study}

In order to study the effects for each of the above unsupervised loss functions and different methods of data processing, we designed ablation studies. Tables ~\ref{table:ablation1},~\ref{table:ablation2} show the results of ablation studies. The experiments prove that each of our methods can improve performance.

We started with the evaluation of the effect of each loss function proposed in Section ~\ref{section:unsupervised learning}. As shown in table ~\ref{table:ablation1}, from the drop in performance we can see  the lack of each loss function will do harm to the evaluation result. And in all loss functions, Chamfer loss plays a crucial role. Then we compared the influence of different data processing methods on the performance. From Table~\ref{table:ablation2} we can see the importance of data processing. Proper methods which remove the points of ground and sky improve the performance greatly.

\section{Conclusion}
This paper used depth network and pose network to generate colorized point clouds and put forward the idea of dynamic and static scene flow. Further, it proposed dynamic-static consistency of scene flow and depth consistency of point cloud which can provide an unsupervised learning approach for scene flow. We reduced the cost of the training process greatly as unlabeled data is used in our network training. High accuracy is achieved and the application of scene flow can be supported. 3D odometry estimation method  \cite{wang2020pwclo} based on 2D images using the pseudo  LiDAR obtained from images is our future work.


\bibliographystyle{IEEEtran}  
\bibliography{IEEEabrv,root} 

\begin{thebibliography}{10}
\providecommand{\url}[1]{#1}
\csname url@rmstyle\endcsname
\providecommand{\newblock}{\relax}
\providecommand{\bibinfo}[2]{#2}
\providecommand\BIBentrySTDinterwordspacing{\spaceskip=0pt\relax}
\providecommand\BIBentryALTinterwordstretchfactor{4}
\providecommand\BIBentryALTinterwordspacing{\spaceskip=\fontdimen2\font plus
\BIBentryALTinterwordstretchfactor\fontdimen3\font minus
  \fontdimen4\font\relax}
\providecommand\BIBforeignlanguage[2]{{%
\expandafter\ifx\csname l@#1\endcsname\relax
\typeout{** WARNING: IEEEtran.bst: No hyphenation pattern has been}%
\typeout{** loaded for the language `#1'. Using the pattern for}%
\typeout{** the default language instead.}%
\else
\language=\csname l@#1\endcsname
\fi
#2}}

\bibitem{liu2016formation}
Z.~Liu, W.~Chen, J.~Lu, H.~Wang, and J.~Wang, ``Formation control of mobile
  robots using distributed controller with sampled-data and communication
  delays,'' \emph{IEEE Transactions on Control Systems Technology}, vol.~24,
  no.~6, pp. 2125--2132, 2016.

\bibitem{vedula1999three}
S.~Vedula, S.~Baker, P.~Rander, R.~Collins, and T.~Kanade, ``Three-dimensional
  scene flow,'' in \emph{Proceedings of the Seventh IEEE International
  Conference on Computer Vision}, vol.~2.\hskip 1em plus 0.5em minus
  0.4em\relax IEEE, 1999, pp. 722--729.

\bibitem{zhao2019self}
X.~Zhao and H.~Wang, ``Self-localization using point cloud matching at the
  object level in outdoor environment,'' in \emph{2019 IEEE 9th Annual
  International Conference on CYBER Technology in Automation, Control, and
  Intelligent Systems (CYBER)}.\hskip 1em plus 0.5em minus 0.4em\relax IEEE,
  2019, pp. 1447--1452.

\bibitem{liu2019local}
Y.~Liu, H.~Wang, F.~Xu, Y.~Wang, W.~Chen, and Q.~Tang, ``Local pose
  optimization with an attention-based neural network,'' in \emph{2019 IEEE/RSJ
  International Conference on Intelligent Robots and Systems (IROS)}.\hskip 1em
  plus 0.5em minus 0.4em\relax IEEE, 2019, pp. 3084--3089.

\bibitem{shi2016visual}
Y.~Shi and H.~Wang, ``Visual tracking via an ensemble of random classifiers,''
  in \emph{2016 IEEE International Conference on Real-time Computing and
  Robotics (RCAR)}.\hskip 1em plus 0.5em minus 0.4em\relax IEEE, 2016, pp.
  603--608.

\bibitem{vogel20153d}
C.~Vogel, K.~Schindler, and S.~Roth, ``3d scene flow estimation with a
  piecewise rigid scene model,'' \emph{International Journal of Computer
  Vision}, vol. 115, no.~1, pp. 1--28, 2015.

\bibitem{menze2015object}
M.~Menze and A.~Geiger, ``Object scene flow for autonomous vehicles,'' in
  \emph{Proceedings of the IEEE Conference on Computer Vision and Pattern
  Recognition}, 2015, pp. 3061--3070.

\bibitem{menze2018object}
M.~Menze, C.~Heipke, and A.~Geiger, ``Object scene flow,'' \emph{ISPRS Journal
  of Photogrammetry and Remote Sensing}, vol. 140, pp. 60--76, 2018.

\bibitem{lv2016continuous}
Z.~Lv, C.~Beall, P.~F. Alcantarilla, F.~Li, Z.~Kira, and F.~Dellaert, ``A
  continuous optimization approach for efficient and accurate scene flow,'' in
  \emph{European Conference on Computer Vision}.\hskip 1em plus 0.5em minus
  0.4em\relax Springer, 2016, pp. 757--773.

\bibitem{huguet2007variational}
F.~Huguet and F.~Devernay, ``A variational method for scene flow estimation
  from stereo sequences,'' in \emph{2007 IEEE 11th International Conference on
  Computer Vision}.\hskip 1em plus 0.5em minus 0.4em\relax IEEE, 2007, pp.
  1--7.

\bibitem{hussain2019depth}
H.~Abid and H.~Wang, ``Depth-wise pooling: A parameter-less solution for
  channel reduction of feature-map in convolutional neural network,'' in
  \emph{2019 IEEE International Conference on Real-time Computing and Robotics
  (RCAR)}.\hskip 1em plus 0.5em minus 0.4em\relax IEEE, 2019, pp. 299--304.

\bibitem{wang2019depth}
Y.~Wang, W.~Chen, and H.~Wang, ``Depth estimation and background segmentation
  for deformable packages from a single image using fcrn,'' in \emph{2019 IEEE
  International Conference on Real-time Computing and Robotics (RCAR)}.\hskip
  1em plus 0.5em minus 0.4em\relax IEEE, 2019, pp. 267--272.

\bibitem{mayer2016large}
N.~Mayer, E.~Ilg, P.~Hausser, P.~Fischer, D.~Cremers, A.~Dosovitskiy, and
  T.~Brox, ``A large dataset to train convolutional networks for disparity,
  optical flow, and scene flow estimation,'' in \emph{Proceedings of the IEEE
  Conference on Computer Vision and Pattern Recognition}, 2016, pp. 4040--4048.

\bibitem{lv2018learning}
Z.~Lv, K.~Kim, A.~Troccoli, D.~Sun, J.~M. Rehg, and J.~Kautz, ``Learning
  rigidity in dynamic scenes with a moving camera for 3d motion field
  estimation,'' in \emph{Proceedings of the European Conference on Computer
  Vision (ECCV)}, 2018, pp. 468--484.

\bibitem{jiang2019sense}
H.~Jiang, D.~Sun, V.~Jampani, Z.~Lv, E.~Learned-Miller, and J.~Kautz, ``Sense:
  a shared encoder network for scene-flow estimation,'' in \emph{Proceedings of
  the IEEE International Conference on Computer Vision}, 2019, pp. 3195--3204.

\bibitem{gu2019hplflownet}
X.~Gu, Y.~Wang, C.~Wu, Y.~J. Lee, and P.~Wang, ``Hplflownet: Hierarchical
  permutohedral lattice flownet for scene flow estimation on large-scale point
  clouds,'' in \emph{Proceedings of the IEEE Conference on Computer Vision and
  Pattern Recognition}, 2019, pp. 3254--3263.

\bibitem{liu2019flownet3d}
X.~Liu, C.~R. Qi, and L.~J. Guibas, ``Flownet3d: Learning scene flow in 3d
  point clouds,'' in \emph{Proceedings of the IEEE Conference on Computer
  Vision and Pattern Recognition}, 2019, pp. 529--537.

\bibitem{dosovitskiy2015flownet}
A.~Dosovitskiy, P.~Fischer, E.~Ilg, P.~Hausser, C.~Hazirbas, V.~Golkov, P.~Van
  Der~Smagt, D.~Cremers, and T.~Brox, ``Flownet: Learning optical flow with
  convolutional networks,'' in \emph{Proceedings of the IEEE international
  conference on computer vision}, 2015, pp. 2758--2766.

\bibitem{ren2017unsupervised}
Z.~Ren, J.~Yan, B.~Ni, B.~Liu, X.~Yang, and H.~Zha, ``Unsupervised deep
  learning for optical flow estimation,'' in \emph{Thirty-First AAAI Conference
  on Artificial Intelligence}, 2017.

\bibitem{meister2018unflow}
S.~Meister, J.~Hur, and S.~Roth, ``Unflow: Unsupervised learning of optical
  flow with a bidirectional census loss,'' in \emph{Thirty-Second AAAI
  Conference on Artificial Intelligence}, 2018.

\bibitem{wang2018occlusion}
Y.~Wang, Y.~Yang, Z.~Yang, L.~Zhao, P.~Wang, and W.~Xu, ``Occlusion aware
  unsupervised learning of optical flow,'' in \emph{Proceedings of the IEEE
  Conference on Computer Vision and Pattern Recognition}, 2018, pp. 4884--4893.

\bibitem{janai2018unsupervised}
J.~Janai, F.~Guney, A.~Ranjan, M.~Black, and A.~Geiger, ``Unsupervised learning
  of multi-frame optical flow with occlusions,'' in \emph{Proceedings of the
  European Conference on Computer Vision (ECCV)}, 2018, pp. 690--706.

\bibitem{liu2019ddflow}
P.~Liu, I.~King, M.~R. Lyu, and J.~Xu, ``Ddflow: Learning optical flow with
  unlabeled data distillation,'' \emph{arXiv preprint arXiv:1902.09145}, 2019.

\bibitem{liu2019selflow}
P.~Liu, M.~Lyu, I.~King, and J.~Xu, ``Selflow: Self-supervised learning of
  optical flow,'' in \emph{Proceedings of the IEEE Conference on Computer
  Vision and Pattern Recognition}, 2019, pp. 4571--4580.

\bibitem{geiger2013vision}
A.~Geiger, P.~Lenz, C.~Stiller, and R.~Urtasun, ``Vision meets robotics: The
  kitti dataset,'' \emph{The International Journal of Robotics Research},
  vol.~32, no.~11, pp. 1231--1237, 2013.

\bibitem{behl2017bounding}
A.~Behl, O.~Hosseini~Jafari, S.~Karthik~Mustikovela, H.~Abu~Alhaija, C.~Rother,
  and A.~Geiger, ``Bounding boxes, segmentations and object coordinates: How
  important is recognition for 3d scene flow estimation in autonomous driving
  scenarios?'' in \emph{Proceedings of the IEEE International Conference on
  Computer Vision}, 2017, pp. 2574--2583.

\bibitem{battrawy2019lidar}
R.~Battrawy, R.~Schuster, O.~Wasenm{\"u}ller, Q.~Rao, and D.~Stricker,
  ``Lidar-flow: Dense scene flow estimation from sparse lidar and stereo
  images,'' \emph{arXiv preprint arXiv:1910.14453}, 2019.

\bibitem{ma2019deep}
W.-C. Ma, S.~Wang, R.~Hu, Y.~Xiong, and R.~Urtasun, ``Deep rigid instance scene
  flow,'' in \emph{Proceedings of the IEEE Conference on Computer Vision and
  Pattern Recognition}, 2019, pp. 3614--3622.

\bibitem{dewan2016rigid}
A.~Dewan, T.~Caselitz, G.~D. Tipaldi, and W.~Burgard, ``Rigid scene flow for 3d
  lidar scans,'' in \emph{2016 IEEE/RSJ International Conference on Intelligent
  Robots and Systems (IROS)}.\hskip 1em plus 0.5em minus 0.4em\relax IEEE,
  2016, pp. 1765--1770.

\bibitem{ushani2017learning}
A.~K. Ushani, R.~W. Wolcott, J.~M. Walls, and R.~M. Eustice, ``A learning
  approach for real-time temporal scene flow estimation from lidar data,'' in
  \emph{2017 IEEE International Conference on Robotics and Automation
  (ICRA)}.\hskip 1em plus 0.5em minus 0.4em\relax IEEE, 2017, pp. 5666--5673.

\bibitem{qi2017pointnet}
C.~R. Qi, H.~Su, K.~Mo, and L.~J. Guibas, ``Pointnet: Deep learning on point
  sets for 3d classification and segmentation,'' in \emph{Proceedings of the
  IEEE Conference on Computer Vision and Pattern Recognition}, 2017, pp.
  652--660.

\bibitem{qi2017pointnet++}
C.~R. Qi, L.~Yi, H.~Su, and L.~J. Guibas, ``Pointnet++: Deep hierarchical
  feature learning on point sets in a metric space,'' in \emph{Advances in
  neural information processing systems}, 2017, pp. 5099--5108.

\bibitem{liu2019lpd}
Z.~Liu, S.~Zhou, C.~Suo, P.~Yin, W.~Chen, H.~Wang, H.~Li, and Y.-H. Liu,
  ``Lpd-net: 3d point cloud learning for large-scale place recognition and
  environment analysis,'' in \emph{Proceedings of the IEEE/CVF International
  Conference on Computer Vision}, 2019, pp. 2831--2840.

\bibitem{wang2020spherical}
G.~Wang, Y.~Yang, H.~Zhang, Z.~Liu, and H.~Wang, ``Spherical interpolated
  convolutional network with distance-feature density for 3d semantic
  segmentation of point clouds,'' \emph{arXiv preprint arXiv:2011.13784}, 2020.

\bibitem{wang2020anchor}
G.~Wang, H.~Liu, M.~Chen, Y.~Yang, Z.~Liu, and H.~Wang, ``Anchor-based
  spatial-temporal attention convolutional networks for dynamic 3d point cloud
  sequences,'' \emph{arXiv preprint arXiv:2012.10860}, 2020.

\bibitem{behl2019pointflownet}
A.~Behl, D.~Paschalidou, S.~Donn{\'e}, and A.~Geiger, ``Pointflownet: Learning
  representations for rigid motion estimation from point clouds,'' in
  \emph{Proceedings of the IEEE Conference on Computer Vision and Pattern
  Recognition}, 2019, pp. 7962--7971.

\bibitem{wang2018deep}
S.~Wang, S.~Suo, W.-C. Ma, A.~Pokrovsky, and R.~Urtasun, ``Deep parametric
  continuous convolutional neural networks,'' in \emph{Proceedings of the IEEE
  Conference on Computer Vision and Pattern Recognition}, 2018, pp. 2589--2597.

\bibitem{wu2020pointpwc}
W.~Wu, Z.~Y. Wang, Z.~Li, W.~Liu, and L.~Fuxin, ``Pointpwc-net: Cost volume on
  point clouds for (self-) supervised scene flow estimation,'' in
  \emph{European Conference on Computer Vision}.\hskip 1em plus 0.5em minus
  0.4em\relax Springer, 2020, pp. 88--107.

\bibitem{wang2020hierarchical}
G.~Wang, X.~Wu, Z.~Liu, and H.~Wang, ``Hierarchical attention learning of scene
  flow in 3d point clouds,'' \emph{arXiv preprint arXiv:2010.05762}, 2020.

\bibitem{ilg2017flownet}
E.~Ilg, N.~Mayer, T.~Saikia, M.~Keuper, A.~Dosovitskiy, and T.~Brox, ``Flownet
  2.0: Evolution of optical flow estimation with deep networks,'' in
  \emph{Proceedings of the IEEE conference on computer vision and pattern
  recognition}, 2017, pp. 2462--2470.

\bibitem{besl1992method}
P.~J. Besl and N.~D. McKay, ``Method for registration of 3-d shapes,'' in
  \emph{Sensor fusion IV: control paradigms and data structures}, vol.
  1611.\hskip 1em plus 0.5em minus 0.4em\relax International Society for Optics
  and Photonics, 1992, pp. 586--606.

\bibitem{ilg2018occlusions}
E.~Ilg, T.~Saikia, M.~Keuper, and T.~Brox, ``Occlusions, motion and depth
  boundaries with a generic network for disparity, optical flow or scene flow
  estimation,'' in \emph{Proceedings of the European Conference on Computer
  Vision (ECCV)}, 2018, pp. 614--630.

\bibitem{yin2018geonet}
Z.~Yin and J.~Shi, ``Geonet: Unsupervised learning of dense depth, optical flow
  and camera pose,'' in \emph{Proceedings of the IEEE Conference on Computer
  Vision and Pattern Recognition}, 2018, pp. 1983--1992.

\bibitem{Ranjan2019CCNet}
A.~Ranjan, V.~Jampani, L.~Balles, K.~Kim, D.~Sun, J.~Wulff, and M.~J. Black,
  ``Competitive collaboration: Joint unsupervised learning of depth, camera
  motion, optical flow and motion segmentation,'' in \emph{Proceedings of the
  IEEE conference on computer vision and pattern recognition}.\hskip 1em plus
  0.5em minus 0.4em\relax IEEE, 2019, pp. 12\,240--12\,249.

\bibitem{wang2020unsupervised}
G.~Wang, C.~Zhang, H.~Wang, J.~Wang, Y.~Wang, and X.~Wang, ``Unsupervised
  learning of depth, optical flow and pose with occlusion from 3d geometry,''
  \emph{IEEE Transactions on Intelligent Transportation Systems}, 2020.

\bibitem{wang2019unsupervised}
G.~Wang, H.~Wang, Y.~Liu, and W.~Chen, ``Unsupervised learning of monocular
  depth and ego-motion using multiple masks,'' in \emph{2019 International
  Conference on Robotics and Automation (ICRA)}.\hskip 1em plus 0.5em minus
  0.4em\relax IEEE, 2019, pp. 4724--4730.

\bibitem{hur2020self}
J.~Hur and S.~Roth, ``Self-supervised monocular scene flow estimation,'' in
  \emph{Proceedings of the IEEE/CVF Conference on Computer Vision and Pattern
  Recognition}, 2020, pp. 7396--7405.

\bibitem{bian2019unsupervised}
J.-W. Bian, Z.~Li, N.~Wang, H.~Zhan, C.~Shen, M.-M. Cheng, and I.~Reid,
  ``Unsupervised scale-consistent depth and ego-motion learning from monocular
  video,'' \emph{arXiv preprint arXiv:1908.10553}, 2019.

\bibitem{sorkine2005laplacian}
O.~Sorkine, ``Laplacian mesh processing,'' \emph{Eurographics (STARs)},
  vol.~29, 2005.

\bibitem{zhou2016fast}
Q.-Y. Zhou, J.~Park, and V.~Koltun, ``Fast global registration,'' in
  \emph{European Conference on Computer Vision}.\hskip 1em plus 0.5em minus
  0.4em\relax Springer, 2016, pp. 766--782.

\bibitem{wang2020pwclo}
G.~Wang, X.~Wu, Z.~Liu, and H.~Wang, ``Pwclo-net: Deep lidar odometry in 3d
  point clouds using hierarchical embedding mask optimization,''
  \emph{Proceedings of the IEEE/CVF Conference on Computer Vision and Pattern
  Recognition}, 2021.

\end{thebibliography}

\end{document}